\documentclass[letterpaper, conference, twocolumn, final]{ieeeconf}
\IEEEoverridecommandlockouts 

\usepackage{graphicx} 
\usepackage{booktabs}
\usepackage[dvipsnames]{xcolor}
\usepackage{amsfonts,amsmath,amssymb}
\usepackage[utf8]{inputenc}
\usepackage{epstopdf}
\usepackage{subfigure}
\usepackage{soul}
\usepackage{hyperref}
\graphicspath{{figs/}}
\usepackage{todonotes}

\newcommand{\nbrs}{\mathcal{N}}

\newcommand{\real}{\mathbb{R}}

\newcommand{\dx}{\dot{x}}
\newcommand{\EE}{\mathcal{E}}
\newcommand{\GG}{\mathcal{G}}

\newcommand{\XX}{\mathcal{X}}
\newcommand{\UU}{\mathcal{U}}
\renewcommand{\SS}{\mathcal{S}}
\newcommand{\until}[1]{\{1, \ldots, #1\}}

\usepackage{listings}
\definecolor{mygray}{gray}{0.95}
\usepackage{cuted}
\def \packagename/{\textsc{ChoiRbot}}

\title{\LARGE \bf ChoiRbot: A ROS~2 Toolbox for Cooperative Robotics}

\author{Andrea Testa$^{\dagger}$,
  Andrea Camisa$^{\dagger}$,
  Giuseppe Notarstefano%
\thanks{This work was supported by
the European 
  Research Council (ERC) under the European Union's Horizon 2020 research 
  and innovation programme (grant agreement No 638992 - OPT4SMART).}%
  \thanks{A. Testa, A. Camisa and G. Notarstefano are with the Department of Electrical, 
  Electronic and Information Engineering, University of Bologna, Bologna, Italy. 
  \texttt{\{a.testa, a.camisa, giuseppe.notarstefano\}@unibo.it}.}%
  \thanks{$^{\dagger}$ These authors contributed equally to this work.}%
}
\begin{document}

\maketitle
\begin{strip}\leavevmode\kern15pt
\begin{minipage}{\dimexpr\linewidth-30pt\relax}
{\vspace{-2.5cm}
\bf \textcopyright 2021 IEEE. Personal use of this material is permitted.  Permission from IEEE must be obtained for all other uses, in any current or future media, including reprinting/republishing this material for advertising or promotional purposes, creating new collective works, for resale or redistribution to servers or lists, or reuse of any copyrighted component of this work in other works.}
\end{minipage}
\end{strip}

\begin{abstract} %
  In this paper, we introduce \packagename/, a toolbox for distributed cooperative
  robotics based on the novel Robot Operating System (ROS) 2.
  \packagename/ provides a fully-functional toolset to execute complex distributed
  multi-robot tasks, either in simulation or experimentally, with a particular focus on
  networks of heterogeneous robots without a central coordinator.
  Thanks to its modular structure, \packagename/ allows for a highly
  straight implementation of optimization-based distributed control schemes,
  such as distributed optimal control, model predictive control, task assignment, in which
  local computation and communication with neighboring robots are alternated.
  To this end, the toolbox provides functionalities for the solution of
  distributed optimization problems.
  The package can be also used to implement distributed feedback laws
  that do not need optimization features
  but do require the exchange of information among robots.
  The potential of the toolbox is illustrated with simulations and experiments on
  distributed robotics scenarios with mobile ground robots.
  The \packagename/ toolbox is available at \href{https://github.com/OPT4SMART/choirbot}{\texttt{https://github.com/OPT4SMART/choirbot}}.
\end{abstract}


\section{Introduction}
\label{sec:introduction}
There is an increasing interest towards several applications in cooperative robotics, such as task
assignment/allocation, model predictive control, formation control. In peer-to-peer robotic networks
it is often desirable to solve these problems without a coordinating unit. Robots in the network, relying on
their limited knowledge of the problem data, have to exploit local computation and communication
capabilities to solve the complex task, which often requires also the solution of optimization
problems.
Since its introduction, the Robot Operating System (ROS),~\cite{quigley2009ros}, has gained popularity
among robotics researchers as an open source framework for the development of robotics applications.
Nowadays, ROS 2 is extending ROS capabilities, paving the way to real-time control systems and
large-scale distributed architectures,~\cite{maruyama2016exploring}.
While several theoretical frameworks have been proposed to solve optimization problems over networks
of cooperating robots, see, e.g. the survey~\cite{notarstefano2019distributed} and references therein,
few architectures have been proposed to simulate and run experiments on teams of heterogeneous
robots communicating according to arbitrary graphs and aiming at the cooperative solution of complex tasks.

\subsection{Related Work}
The ROS framework has been used as a building brick for a plethora of robotics applications,
and several frameworks have been proposed to simulate and implement control and
planning tasks.
Authors in~\cite{aertbelien2014etasl} propose a constraint-based task specification for robot
controllers, defining a task specification language and the related controller.
The framework in~\cite{paxton2017costar} allows users to create task plans for collaborative robots.
Papers~\cite{grabe2013telekyb} and~\cite{meyer2012comprehensive} instead propose architectures
to simulate and control UAVs, while~\cite{casan2015ros} allows users to write ROS code on a
browser and run it on remote robots. On this purpose, it is worth mentioning the Robotarium,
\cite{wilson2020robotarium}, a proprietary platform that allows to test and run control
algorithms on robotic teams. These frameworks often leverage optimization routines to perform
specific tasks. To name a few, the framework in~\cite{albers2019online} deals with trajectory
optimization in dynamic environments, the architecture in~\cite{kumar2017search} approaches
search-and-rescue tasks by means of particle swarm optimization while~\cite{dos2016implementing}
is suited for task allocation scenarios.
Finally, in the recent years, robotics researchers started to develop robotic architectures based on the
novel ROS 2 framework. Authors in~\cite{erHos2019ros2} and~\cite{erHos2019integrated} consider a
framework for collaborative robotics, while the paper in~\cite{reke2020self} discusses an
architecture for self-driving cars.
The aforementioned frameworks are typically optimized for a specific task, and most of
them focus on single-robot systems. Moreover, the communication in multi-robot networks
is often neglected or simulated by means of the resource-demanding all-to-all communication.

\subsection{Contributions}
In this paper, we introduce a novel ROS 2 toolbox for cooperative robotics named
\packagename/. This toolbox, written in Python, exploits the new functionalities of ROS 2
and provides a comprehensive set of libraries to facilitate multi-robot simulations and experiments.
The main focus of \packagename/ is on peer-to-peer network of robots,
where each robot has its own processor and is able to communicate with the neighboring
units according to a user-defined graph, possibly time-varying or with unreliable
communication links. Importantly, \packagename/ does not
require a central coordinating unit and, as such, allows for fully distributed control schemes.
In order to maximize extendability and ensure a broad applicability, the toolbox is designed
according to a modular structure. In such a way, we demonstrate that several applications
of interest are easy to implement using \packagename/, and we discuss in detail a few
use cases that have been tested either in simulation or experimentally, namely
dynamic task assignment, formation control and containment in leader-follower networks.
A distinctive feature of \packagename/ is the ability to run general-purpose
distributed optimization algorithms. To this end, the \textsc{DISROPT}
package~\cite{farina2019disropt} has been fully integrated, allowing
both for the semantic modeling and solution of optimization problems (locally at a robot)
and for the execution of distributed optimization algorithms (cooperatively across all the robots).

The paper is organized as follows. In Section~\ref{sec:architecture_description}, we provide
a high-level description of the software architecture, while in Section~\ref{sec:guidance_details},
additional software details are provided. Section~\ref{sec:complex_scenarios} discusses two
complex scenarios implemented in \packagename/. A basic use case together with implementation
details is discussed in~\ref{sec:basic_usage}. Finally, simulation and experimental results
are provided in Section~\ref{sec:simexp}.

\section{Architecture Description}
\label{sec:architecture_description}
In this section, we describe the high-level architecture of \packagename/. As already mentioned,
the toolbox is modular and its blocks are intended to be combined as needed.
We first focus on an overall description of the software and then we
describe each block separately.

\subsection{Overview of the Software}
\label{sec:architecture_overview}
\packagename/ is written in Python and is based on
ROS~2.
An important feature of the new ROS 2 architecture is that
no master entity is present~\cite{maruyama2016exploring}.
Therefore, \textsc{ChoiRbot} processes are truly distributed since
no message broker is required.
This fits perfectly with the goal of \packagename/ of providing a
platform for cooperative robotics over peer-to-peer networks without a
central coordinating unit.

The toolbox is structured in a three-layer architecture.
Specifically, there is a \emph{Team Guidance} layer, a
\emph{RoboPlanning} layer and a \emph{RoboControl} layer.
The Team Guidance layer is responsible for taking high-level decisions and for
managing the robot lifecycle. %
The Team Guidance layer uses communication with neighbors in order to perform its tasks.
The Roboplanning and Robocontrol layers are responsible for lower-level control actions as
driven by the upper layer.
Since our main goal is to ease the design and implementation of
optimization-based distributed control schemes, the central focus of
\packagename/ is the Team Guidance layer. In particular, the toolbox provides
boilerplate code for distributed computation, i.e. communication
among robots over a graph, coordination and optimization algorithms.
Apart from a few specific scenarios, we do not provide comprehensive RoboPlanning
and RoboControl features, which are robot specific and can be already implemented with
existing tools.

\begin{figure*}[t]\centering
\vspace{0.25cm}
  \includegraphics[scale=0.8]{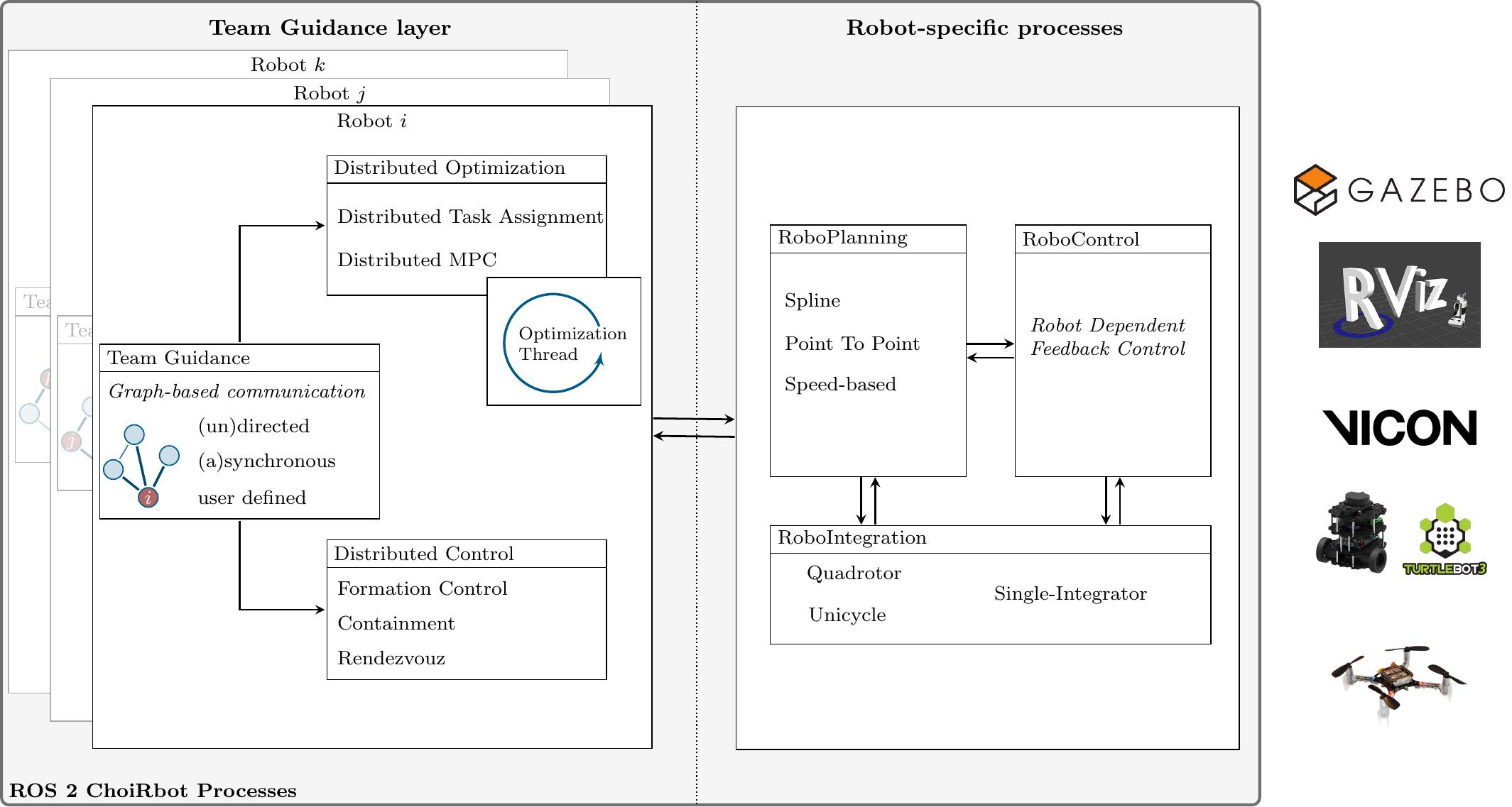}
\caption{\packagename/ architecture. Each robot communicates with neighboring robotic agents thanks to the Team Guidance Layer. The Team Guidance class handles the communication with planning and control utilities, which are specific for each robot.}
\label{fig:architecture}
\end{figure*}

In \packagename/, each robotic agent in the network executes three ROS 2
processes, one for each layer. Each process is also associated to a separate ROS 2 node.
To guarantee flexibility and code reusability, layers are implemented as Python classes.
In the remainder of this section, we provide details about the three layers.
A graphical illustration of the software architecture is represented in Figure~\ref{fig:architecture}.

\subsection{Team Guidance Layer}
\label{sec:team_guidance_description_short}
The Team Guidance layer is the main entry point of the package. Here we provide
an introductory description, while a more detailed analysis of this part of software
is delayed to Section~\ref{sec:guidance_details}.

The Team Guidance layer is modeled with a \texttt{Guidance} class, whose purpose
is only to retrieve the basic information regarding the robotic agent
(passed as ROS 2 parameters), to create an instance of the \texttt{Communicator}
class to be described in Section~\ref{sec:communicator} and to subscribe to
the topic where the robot pose is published.
In the current version of \packagename/, the robot position can be either
communicated by a Vicon motion tracking system or by a simulator (such as Gazebo).
This basic version of the \texttt{Guidance} class is abstract in that it does not implement any
guidance logic. However, the package currently provides two possible usable extensions.
The first one (implemented as the \texttt{OptimizationGuidance} class)
also provides optimization-related functionalities and is the starting point
for any optimization-based distributed control scheme such as task assignment / allocation,
optimal control, model predictive control.
The toolbox can also be used to implement simpler distributed feedback
laws that do not require the solution of optimization problems. To this end, we also provide
a second extension of the \texttt{Guidance} class (implemented as the
\texttt{DistributedControlGuidance} class), where robots repeatedly exchange
their current position with their neighbors and compute a control input based
on their position and the received positions. The actual form of the control input
depends on the specific scenario and is left as an unimplemented method of the
class. Currently implemented algorithms are rendezvous, containment, formation control.

\subsection{RoboPlanning, RoboControl and RoboIntegration}
\label{sec:planning_control_integration_description}
Within the \packagename/ toolbox, the trajectory planning layer and
the control layer do not communicate with the neighboring robotic agents.
Instead, each robot has its own planning/control stack, which depends on
the dynamics of the single robot and on the chosen control strategy.
The base class for trajectory planners is \texttt{Planner}.
We provide a point-to-point planner (in 2D or 3D), to be used in
conjunction with controllers that are able to steer the robot towards
a specified point in the space (e.g. for mobile ground robots).
We also provide trajectory planners for aerial robots, which generate quad-rotor
trajectories either to reach a desired constant speed or to reach a target point.
As for controllers, the base class is \texttt{Controller}. For mobile ground
robots, we provide two unicycle controllers that can be used either to
reach a target point or to reach a desired velocity. For quadrotors, we
provide a controller to stabilize the trajectories generated by the planners.

The \packagename/ toolbox is well integrated with simulation environments.
In this regard, the \texttt{Gazebo} simulator \cite{koenig2004design} can be used.
In case the user does not want to use external tools, we also provide a
 dynamics integration layer, named \emph{RoboIntegration}, which can e.g. be used in conjunction
with \texttt{Rviz} for visualization.
Currently, there are numerical integrators for point-masses, unicycle robots
and quadrotors. New integrators with custom dynamics can be written
by extending the \texttt{Integrator} class.

\section{Exploring the Team Guidance Layer}
\label{sec:guidance_details}
As already mentioned, almost all functionalities of \packagename/ involve
the Team Guidance layer. In this section, we explore in more detail this part of software.
We begin by analyzing the first important feature of the Team Guidance class, i.e.
graph-based communication.
Then, we analyze the two essential usages of the Team Guidance layer, represented
by the \texttt{OptimizationGuidance} and the \texttt{DistributedControlGuidance} classes.

\subsection{Graph-based Communication}
\label{sec:communicator}
At the core of every distributed control scheme is the graph-based communication
among the robotic agents. This feature is provided in \packagename/ at the team
guidance layer by three different \texttt{Communicator} classes,
which model static, time-varying and unreliable (``best-effort'')
communication, respectively.
By exploiting the novel Quality of Service introduced in ROS~2,
these classes can handle synchronous/asynchronous and
undirected/directed communication among the robots, as summarized
in Table~\ref{tab:communicators}.
To achieve this, the \texttt{Communicator} class only requires specification
of the in- and out-neighbors of the robots and
takes care of managing the necessary ROS 2 topics and subscriptions. 
To maintain the class interface of the class semantically clear, the method
names correspond to the specific actions that can be performed: \texttt{send},
\texttt{receive}, \texttt{asynchronous\_receive}, \texttt{neighbors\_exchange}
(i.e. simultaneous \texttt{send}/\texttt{receive}).
Depending on the specific application, the \texttt{Communicator}
classes can be used to handle scenarios in which the communication links
change on the fly due to, e.g., limited communication range and/or energy
consumption constraints.

\setlength\tabcolsep{2.5pt}
\begin{table}[htpb]\centering\scriptsize
\begin{tabular}{lcccccccc}
\toprule
\bf Class name & \bf \begin{tabular}{c}Directed \\ Undirected \end{tabular} & \bf \begin{tabular}{c}Time \\ Varying\end{tabular} & \bf Synch. & \bf Asynch. & \bf Reliable \\
\midrule
\texttt{StaticComm.} & \checkmark & & \checkmark & \checkmark & \checkmark & \\
\texttt{TimeVaryingComm.} & \checkmark & \checkmark & \checkmark & \checkmark & \checkmark \\
\texttt{BestEffortComm.} & \checkmark & \checkmark & & \checkmark & \\
\bottomrule
\end{tabular}
\caption{Features and limitations of the implemented communicators.}
\label{tab:communicators}
\end{table}

An important feature is that it is not necessary to declare the message types
to be exchanged among the robots, which is typically needed for all ROS applications.
Instead, we fixed the type of message to \texttt{std\_msgs/ByteMultiArray}
and the \texttt{Communicator} class handles (de)serialization of the exchanged message
to (from) a byte sequence through the Python \texttt{dill} package.
This allows the user to exchange nearly each type of message
(vector, matrices, text, dictionaries, images, etc.) and even to change it at
runtime without declaring what type of message must be sent.

\subsection{Distributed Optimization}
\label{sec:distributed_optimization}
A consistent part of the toolbox is devoted to providing optimization-related
functions. The main entry point is the \texttt{OptimizationGuidance} class.
Since numerical optimization algorithms may require a certain number of
iterations to converge, these computations are delegated to a separate thread
running in the \texttt{OptimizationThread} class. This allows the ROS 2 guidance
process to continue elaborating callbacks even though an optimization is in progress.
This separate thread class, which is started by the \texttt{OptimizationGuidance} class,
allows the user to start/stop the optimization processes on demand.
At the end of an optimization process, the method \texttt{optimization\_ended} is called.
This method is supposed to be overridden by the user with the specific guidance logic,
e.g. by retrieving the optimization results and by using them as needed.
Depending on the scenario at hand, it may be required that
the problem data is re-evaluated in order to avoid outdated information.
The actual body of the optimization algorithm is left unimplemented so
as to allow the user to implement the desired method. Currently, we provide
implementations for distributed task assignment and distributed
Model Predictive Control scenarios, which are analyzed in detail
in Section~\ref{sec:complex_scenarios}.

The framework allows one to model and to solve both local optimization
problems (at a robot) and distributed optimization problems
(involving the whole network).
To this end, the Team
Guidance layer has been integrated with the \textsc{DISROPT}
package~\cite{farina2019disropt}, which provides a large number
of already implemented distributed optimization schemes
and allows for the semantic modeling of (distributed) optimization problems.
Since the \texttt{Communicator} classes of \textsc{ChoiRbot}
are fully compatible with \textsc{DISROPT}, the distributed algorithms
implemented in \textsc{DISROPT} can be used seamlessly, i.e., with no
further modifications.

\subsection{Distributed Feedback Control}
\label{sec:distributed_feedback_control}
The toolbox is designed to support optimization-based
distributed control algorithms, however simpler distributed feedback laws
can also be implemented by using a subset of the toolbox features.
The \texttt{DistributedControlGuidance} class implements a general communication and
control structure allowing for the implementation of such feedback laws.
To give an idea, let us report an excerpt of the main routine of the class,
which is executed with a user-chosen frequency:
\begin{lstlisting} 
# exchange current position with neighbors
data = self.communicator.neighbors_exchange(
			self.current_pose.position, 
			self.in_neighbors, 
			self.out_neighbors, False)

# compute input
u = self.evaluate_velocity(data)

# send input to planner/controller
self.send_input(u)
\end{lstlisting}
In words, the class first exchanges the current position with the neighbors,
then computes a velocity profile according to the exchanged data and to the
considered feedback law, and finally publishes the control input on a topic.
Despite its simplicity, this basic structure can be used for several distributed
control algorithms such as containment and formation control. Simulation results
can be found in Section~\ref{sec:simexp}.

\section{Basic Usage Example}
\label{sec:basic_usage}
In this section, we consider a toy example that allows us to show the implementation
of a basic distributed cooperative robotics scenario in \packagename/.
Specifically, we consider containment in leader-follower networks as 
described in~\cite{notarstefano2011containment}.
The mathematical formulation is as follows.
Robots communicate according to a time-varying undirected graph
$\GG^t = (V, \EE^t)$, where $V = \until{N}$ is the set of vertices and
$\EE^t \subseteq V \times V$ is the set of edges at time $t$.
We denote by $\nbrs_i^t$ the set of neighbors of each robot $i$
at time $t$.
Robots are partitioned in two groups, namely leaders and followers.
The goal for the followers is to converge to the convex hull
of the leaders' positions. To this end, the robots implement the dynamics
\begin{subequations}
\label{eq:containment}
\begin{align}
  \dx_i(t) &= 0 \hspace{4cm} \text{(leaders)},
  \\
  \dx_i(t) &= \sum_{j \in \nbrs_i^t} \big(x_j(t) - x_i(t) \big) \hspace{1cm} \text{(followers)}.
\end{align}%
\end{subequations}%
\subsection{Main Software Components}
This scenario can be easily implemented in the proposed architecture by means of two
classes in the Team Guidance Layer and in the RoboIntegration layer.
The first one is an extension of the \texttt{DistributedControlGuidance} class discussed in
Section~\ref{sec:distributed_feedback_control} and inherits the distributed feedback control
structure. We simply override
the \texttt{evaluate\_velocity} method (cf. Section~\ref{sec:distributed_feedback_control})
in order to encode the control law
in~\eqref{eq:containment} as follows:
\begin{lstlisting} 
u = np.zeros(3)
if not self.is_leader:
    for pos_ii in neigh_data.values():
        u += self.containment_gain*(pos_ii - self.current_pose.position)
return u
\end{lstlisting}
The second class instead extends the functionalities of the base
class \texttt{Integrator}.
In particular, it is sufficient to %
implement the classical forward Euler method for single-integrator dynamics:
\begin{lstlisting} 
self.current_pos += self.samp_time * self.u
\end{lstlisting}

\subsection{Visualizing the Evolution}
Since we are not using an external simulation tool, we need a means to visualize the
results graphically. To achieve this, it is possible to rely on the ROS 2 toolbox
\textsc{Rviz}, in which we associate each robot with a circle moving on the $(x,y)$ plane.
To achieve this, the toolbox provides a class that receives the pose from
the integration layer and forwards it to the
\textsc{Rviz} visualizer with a user-defined frequency.

\subsection{Running the Simulation}
The final simulation can be run by writing a ROS 2 launch file with all the required nodes.
Note that these three nodes must be executed by each of the robotic agents in the network.
The launcher file is also responsible for declaring the initial positions of the robots
and the communication graph, specified as a binary adjacency matrix.

In order to differentiate the ROS~2 nodes of each robot, robot-specific parameters must be
passed as ROS~2 parameters at the time of spawning. For instance, the Team Guidance classes are spawned in the launch file as follows:
\begin{lstlisting}
def generate_launch_description():
  # .. initialization of adjacency matrix and initial positions
  
  list_description = []
  for i in range(N):
    # .. initialize in_neighbors, out_neighbors, is_leader
    
    list_description.append(Node(
      package='choirbot_examples', node_executable='choirbot_containment',
      node_namespace='agent_{}'.format(i),
      parameters=[{'agent_id': i, 'N': N, 'in_neigh': in_neighbors, 'out_neigh': out_neighbors, 'is_leader': is_leader}]))

  return LaunchDescription(list_description)
\end{lstlisting}

\begin{figure}[ht]
\centering
	\includegraphics[scale=1]{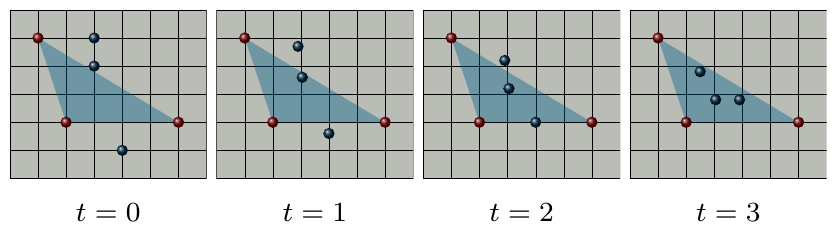}
	\caption{Sequence of images from the \textsc{Rviz} toolbox at different subsequent time instants.
	Red spheres represent leader robots, while the blue ones represent the followers.
	As time progresses, followers enter the convex hull of the leaders' positions,
	depicted with the cerulean triangle.}
	\label{fig:containment}
\end{figure}

A simulation with e.g. $N = 6$ robots can be run by executing the following command:
\begin{lstlisting}
ros2 launch choirbot_examples containment.launch.py -n 6
\end{lstlisting}

We consider a scenario with three followers and three leaders
spanning a triangle.
In order to save energy, at each time instant robots decide whether
to communicate their position to neighbors according
to a certain probability, thus the neighbors change at each
communication round.
The evolution of the robot positions is reported in Figure~\ref{fig:containment}.

\section{Implemented Complex Scenarios}
\label{sec:complex_scenarios}
In this section, we discuss in detail two complex cooperative
robotic scenarios that have been implemented in \packagename/,
i.e. distributed dynamic task assignment and
distributed model predictive control. In both scenarios we employ the
\texttt{OptimizationGuidance} class to leverage optimization capabilities.

\subsection{Distributed Dynamic Task Assignment}
\label{sec:dynamic_task_assignment}
\paragraph*{Problem formulation}
In this scenario, we assume there is a set of tasks that must be performed by
a team of robots. In order to choose the final assignment, robots must self-coordinate
by using their communication capabilities. The assignment problem can be formulated
as an integer optimization problem with unimodularity properties, which is eventually
reformulated into a distributed linear program~\cite{burger2012distributed}.
Formally, assume there are $N$ robots (indexed by $i$) and $N$ tasks (indexed by $k$).
A scalar $c_{ik}$ represents the cost incurred by robot $i$ when servicing task $k$.
The goal is to find the optimal assignment, i.e. to assign each robot $i$ to exactly one task $j$
such that the total incurred cost is minimized.
We focus on the challenging scenario in which tasks are not known a-priori but arrive
dynamically. While robots perform the previously assigned tasks, a new task can arrive and
robots re-execute the distributed optimization algorithm to compute the new optimal assignment.

\paragraph*{Implementation}
In \packagename/, there is a number of classes to handle the dynamic task assignment
scenario. We assume there is a cloud accepting task requests and forwarding them to the
robots such as, e.g. in online order and delivery services, which
is also responsible for maintaining an up-to-date
task list and to mark tasks as completed.
An instance of the \texttt{OptimizationGuidance} class (cf. Section~\ref{sec:distributed_optimization})
communicates with the cloud, %
triggers
the distributed optimization algorithm in the separate computation thread whenever a new
task request is received and maintains a local task queue. While the queue is not empty, tasks
are executed in order %
with the
specific task execution logic (e.g. move to a given point and to perform loading/unloading
operations).
A specific \texttt{OptimizationThread}
is responsible for formulating the
actual linear programming task assignment problem and to run the distributed
simplex algorithm~\cite{burger2012distributed} in \textsc{DISROPT}.
The \texttt{OptimizationThread} class also handle communication among neighboring robots.

Remarkably, from the point of view of the final user, all that is needed is to implement the
task execution logic %
and to integrate %
the cloud with the
task request service of the specific application scenario. The whole dynamic task assignment
mechanism is taken care of by \packagename/.

\subsection{Distributed Model Predictive Control (MPC)}
\label{sec:distributed_mpc}
Let us now describe the second complex scenario implemented in \packagename/.

\paragraph*{Problem formulation}
In this scenario, we assume there are $N$ robots with linear dynamics of the type
$x_i(t+1) = A_i x_i(t) + B_i u_i(t)$, where, for all $i \in \until{N}$,
$x_i(t) \in \real^{n_i}$ and $u_i(t) \in \real^{m_i}$ are the $i$-th robot state/input.
We assume the robots must satisfy local state/input constraints
$x_i(t) \in \XX_i$ and $u_i(t) \in \UU_i$.
Each robot is associated to an output $z_i(t) = C_i x_i(t) + D_i u_i(t)$,
with $z_i \in \real^{r_i}$. The robot outputs are assumed to be coupled with
a constraint $\sum_{i=1}^N z_i(t) \in \SS$.
Finally, we assume each robot $i$ is equipped with a local cost function
$\ell_i(x_i, u_i)$ that must be minimized. The conceptual idea of MPC is to solve a finite-horizon optimal control problem at each time step $t$,
with trajectories spanning the time horizon $[t, t+T]$. The first control input is
applied and the process is repeated. In distributed model predictive control schemes,
each robot solves a local version of the overall optimal control problem and leverages
communication with neighbors to achieve a feasible solution. %

\paragraph*{Implementation}
We implemented the classical distributed MPC algorithm in~\cite{richards2007robust}.
The departing point is a targeted extension of the \texttt{OptimizationGuidance} class, 
which implements the actual steps of the
distributed MPC algorithm. The problem data (i.e. system matrices $A_i$, $B_i$, $C_i$, $D_i$, local
constraints $\XX_i$, $\UU_i$, coupling constraints $\SS$, prediction horizon $T$) are provided
by the user as class parameters.
This class interacts with an \texttt{OptimizationThread} instance
to formulate and solve the local optimal control problem at each
control iteration.
Note that, differently from the dynamic task assignment, in this scenario the solution of
optimization problems is completely local. Here, communication among neighbors occurs within
the \texttt{OptimizationGuidance} class, as required by the MPC algorithm (see~\cite{richards2007robust}
for details).

\section{Simulations and Experiments}
\label{sec:simexp}
In this section, we provide simulation and experimental results for two
different scenarios that can be handled by \packagename/.
We begin by showing experimental results of the dynamic task assignment
scenario described in Section~\ref{sec:dynamic_task_assignment}.
Then, we show how to the proposed package can be easily 
interfaced with Gazebo 
and show simulation results for
a team of mobile wheeled robots.

\subsection{Dynamic Task Assignment on Turtlebot3 Mobile Robots}
\label{sec:task_assignment_experiment}
\begin{figure}[!h]
\centering
  \vspace{-0.2cm}
	\includegraphics{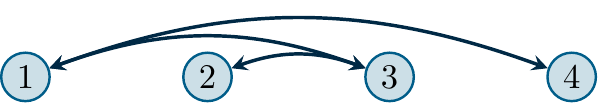} %
	\caption{Communication graph for the dynamic task assignment scenario.}
	\label{fig:taskgraph}
\end{figure}

In this experiment, we consider a team of four Turtlebot 3 Burger mobile
robots that have to accomplish a set of task scattered in the environment.
A task is considered accomplished if the designed robot reaches its position
on the $\{x,y\}$ plane.
As in real applications, problem data are not completely known a-priori, we
consider a dynamic assignment problem where new data arrive during the execution.
Thus, robots have to re-optimize and adjust their local planning whenever new information is available.
Inspired by the approach in~\cite{chopra2017distributed}, 
we assume that a new task is revealed as soon as one has been completed.
The pose of each robot is retrieved by communicating with a Vicon Motion capture system.
In order to interface \packagename/ with the Vicon system, we developed an ad-hoc ROS 2
package, which is provided in the toolbox repository.
As for the control layer to steer robots to the desired positions, we implemented a
dedicated controller node executing the control law in~\cite{park2011smooth}.
Each robot is assigned a fixed label in $\until{4}$.
For the sake of presentation, we assume robots communicate
according to a fixed Erd\H{o}s-R\'{e}nyi graph with edge probability $0.2$,
depicted in Figure~\ref{fig:taskgraph}.
However, the distributed simplex works
under time-varying, asynchronous networks
and is robust to packet losses~\cite{burger2012distributed}.
Thus, it can be also implemented with the \texttt{BestEffortCommunicator}
(cf. Table~\ref{tab:communicators}).

\begin{figure}[ht]
\centering
	\includegraphics[width=.98\columnwidth]{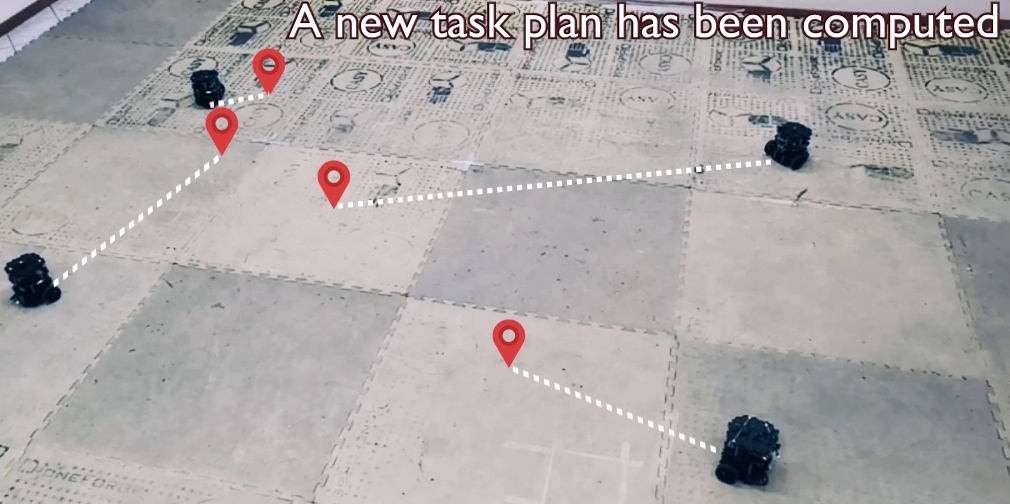}
	\caption{Snapshot from the dynamic task assignment experiment.
	Robots move in order to reach their designed tasks (red markers).}
	\label{fig:exp}
\end{figure}

\begin{figure}[ht]
\centering
  \vspace{0.25cm}
	\includegraphics[width=.98\columnwidth]{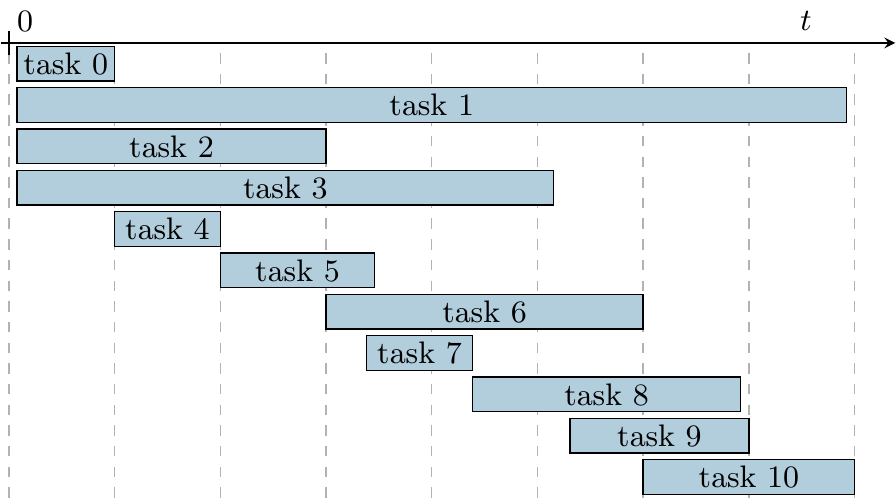}
	\caption{Gantt chart of the task execution flow in the dynamic task
	  assignment. The horizontal axis represents time with the actual
	  scale. The left side of each rectangle indicates the beginning of a task, while the right indicates that the task has been serviced by a robot. As described above, when a task is serviced (e.g. task $0$) a new one begins (e.g. task $4$).}
	\label{fig:gantt}
\end{figure}

When the cloud communicates to robots the pending tasks,
robots start the distributed simplex as described in
Section~\ref{sec:dynamic_task_assignment}.
At each communication round of the distributed algorithm, the
generic robot exchanges with its neighbors a matrix representing
possible robot-to-task assignments until they converge
to a consensual solution. %
In Figure~\ref{fig:exp}, we report a snapshot from the experiment in which
the robots are servicing a set of tasks, while Figure~\ref{fig:gantt} reports a Gantt
chart of the actual task execution.
A video of the experiment is available as supplementary material to the manuscript.%
\footnote{The video is also available at~\url{https://youtu.be/uii1BjFGqMM}.}

\vspace{-0.8cm}

\subsection{Formation Control for Unicycle-Like Robots}
\label{sec:formation_control_simulation}
\begin{figure}[ht]
\centering
  \vspace{0.25cm}
	\includegraphics[width=.99\columnwidth]{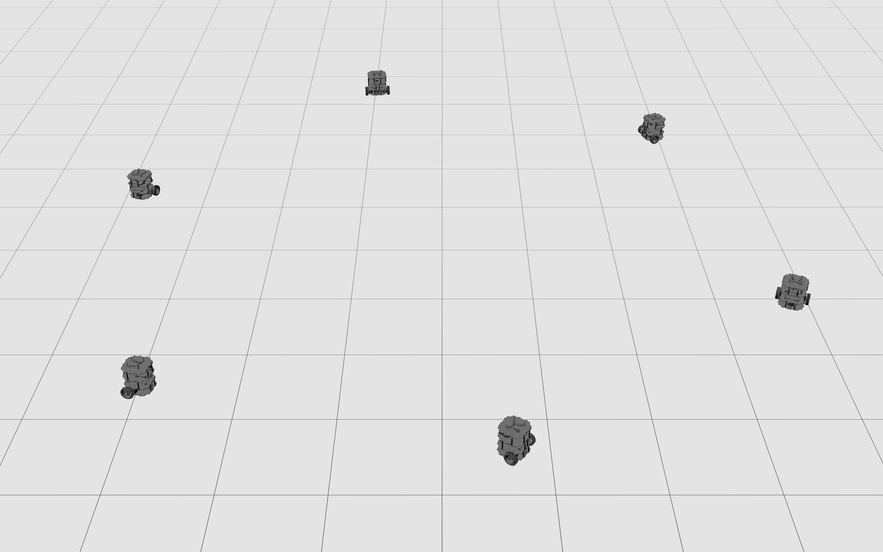}
	\caption{Formation control simulation via Gazebo. By leveraging on \packagename/ functionalities, robots reach the desired hexagonal formation. }
	\label{fig:simgaz}
\end{figure}

In this scenario, the goal is to drive robots to a translationally independent
formation in the $\{x,y\}$ plane that satisfies a set of given constraints.
These constraints are specified by a set of desired inter-robot distances
$d_{ij} \ge 0$ for certain couples $(i,j)$ of robots.
In order to achieve this formation in a distributed way,
robots are assumed to communicate according to a fixed undirected
graph. %
Specifically, all the robot couples $(i,j)$ for which there is a desired
distance $d_{ij}$ are assumed to communicate with each other.
In this scenario, at each control iteration the $i$-th robot communicates
to neighboring robots a vector in $\real^2$ representing its position in the plane.
We refer the reader to, e.g.~\cite{mesbahi2010graph} for a detailed description.
Let the robots apply the distributed control law
\begin{align}\label{eq:formation}
  u_i(t) = \!\sum_{j \in \nbrs_i} \!( \|x_i(t)\! -\! x_j(t)\|^2 - d_{ij}^2) (x_j(t) - x_i(t)),
\end{align}
where $\nbrs_i$ denotes the neighbor set of robot $i$.
We extend the \texttt{DistributedControlGuidance} and
override the \texttt{evaluate\_velocity} method (cf. Section~\ref{sec:distributed_feedback_control})
in order to implement the control law
in~\eqref{eq:formation}. However, the single integrator control input in~\eqref{eq:formation}
 is not directly implementable on unicycle-like robots.
Thus, we developed a specific controller
that maps the input 
provided by~\eqref{eq:formation} to a suitable set of inputs 
for wheeled robots using %
the approach
described in~\cite{wilson2020robotarium}.
The interfacing with Gazebo is straightforward.
Robot poses are retrieved directly by the odometry topic maintained by Gazebo,
while robot inputs are sent on suitable topics read by Gazebo plugins.
In Figure~\ref{fig:simgaz} a snapshot from a Gazebo simulation in which six
Turtlebot 3 Burger robots have to draw an hexagon.
A video of the simulation is available as supplementary material.%
\footnote{The video is also available at~\url{https://youtu.be/VIIXpzPTfPU}.}

\section{Conclusions}
\label{sec:Conclusions}
In this paper, we presented \packagename/, a novel toolbox for distributed cooperative robotics
written in Python and based on the ROS 2 platform. The toolbox is designed with a three-layer structure
and provides a comprehensive set of functionalities for communication and distributed optimization.
Thanks to this toolbox, complex optimization-based cooperative robotic scenarios can be
implemented in a straightforward manner, thus allowing programmers to focus on the most important
part of the production code. Several scenarios have been described and simulations and experimental
results have been presented.
Future directions include the implementation of additional features for the
RoboPlanning/RoboControl layers in order to support additional types
of robots and distributed algorithms.

\bibliographystyle{IEEEtran}
\bibliography{biblio_choirbot}

\begin{thebibliography}{10}
\providecommand{\url}[1]{#1}
\csname url@samestyle\endcsname
\providecommand{\newblock}{\relax}
\providecommand{\bibinfo}[2]{#2}
\providecommand{\BIBentrySTDinterwordspacing}{\spaceskip=0pt\relax}
\providecommand{\BIBentryALTinterwordstretchfactor}{4}
\providecommand{\BIBentryALTinterwordspacing}{\spaceskip=\fontdimen2\font plus
\BIBentryALTinterwordstretchfactor\fontdimen3\font minus
  \fontdimen4\font\relax}
\providecommand{\BIBforeignlanguage}[2]{{%
\expandafter\ifx\csname l@#1\endcsname\relax
\typeout{** WARNING: IEEEtran.bst: No hyphenation pattern has been}%
\typeout{** loaded for the language `#1'. Using the pattern for}%
\typeout{** the default language instead.}%
\else
\language=\csname l@#1\endcsname
\fi
#2}}
\providecommand{\BIBdecl}{\relax}
\BIBdecl

\bibitem{quigley2009ros}
M.~Quigley, K.~Conley, B.~Gerkey, J.~Faust, T.~Foote, J.~Leibs, R.~Wheeler, and
  A.~Y. Ng, ``{ROS}: an open-source {R}obot {O}perating {S}ystem,'' in
  \emph{ICRA workshop on open source software}, vol.~3, no. 3.2.\hskip 1em plus
  0.5em minus 0.4em\relax Kobe, Japan, 2009, p.~5.

\bibitem{maruyama2016exploring}
Y.~Maruyama, S.~Kato, and T.~Azumi, ``Exploring the performance of {ROS}2,'' in
  \emph{Proceedings of the 13th International Conference on Embedded Software},
  2016, pp. 1--10.

\bibitem{notarstefano2019distributed}
\BIBentryALTinterwordspacing
G.~Notarstefano, I.~Notarnicola, and A.~Camisa, ``Distributed optimization for
  smart cyber-physical networks,'' \emph{Foundations and Trends® in Systems
  and Control}, vol.~7, no.~3, pp. 253--383, 2019. [Online]. Available:
  \url{http://dx.doi.org/10.1561/2600000020}
\BIBentrySTDinterwordspacing

\bibitem{aertbelien2014etasl}
E.~Aertbeli{\"e}n and J.~De~Schutter, ``e{T}a{SL}/e{TC}: A constraint-based
  task specification language and robot controller using expression graphs,''
  in \emph{2014 IEEE/RSJ International Conference on Intelligent Robots and
  Systems}.\hskip 1em plus 0.5em minus 0.4em\relax IEEE, 2014, pp. 1540--1546.

\bibitem{paxton2017costar}
C.~Paxton, A.~Hundt, F.~Jonathan, K.~Guerin, and G.~D. Hager, ``Co{STAR}:
  Instructing collaborative robots with behavior trees and vision,'' in
  \emph{2017 IEEE international conference on robotics and automation
  (ICRA)}.\hskip 1em plus 0.5em minus 0.4em\relax IEEE, 2017, pp. 564--571.

\bibitem{grabe2013telekyb}
V.~Grabe, M.~Riedel, H.~H. B{\"u}lthoff, P.~R. Giordano, and A.~Franchi, ``The
  {T}ele{K}yb framework for a modular and extendible {ROS}-based quadrotor
  control,'' in \emph{2013 European Conference on Mobile Robots}.\hskip 1em
  plus 0.5em minus 0.4em\relax IEEE, 2013, pp. 19--25.

\bibitem{meyer2012comprehensive}
J.~Meyer, A.~Sendobry, S.~Kohlbrecher, U.~Klingauf, and O.~Von~Stryk,
  ``Comprehensive simulation of quadrotor uavs using {ROS} and gazebo,'' in
  \emph{International conference on simulation, modeling, and programming for
  autonomous robots}.\hskip 1em plus 0.5em minus 0.4em\relax Springer, 2012,
  pp. 400--411.

\bibitem{casan2015ros}
G.~A. Casan, E.~Cervera, A.~A. Moughlbay, J.~Alemany, and P.~Martinet,
  ``{ROS}-based online robot programming for remote education and training,''
  in \emph{2015 IEEE International Conference on Robotics and Automation
  (ICRA)}.\hskip 1em plus 0.5em minus 0.4em\relax IEEE, 2015, pp. 6101--6106.

\bibitem{wilson2020robotarium}
S.~Wilson, P.~Glotfelter, L.~Wang, S.~Mayya, G.~Notomista, M.~Mote, and
  M.~Egerstedt, ``The robotarium: Globally impactful opportunities, challenges,
  and lessons learned in remote-access, distributed control of multirobot
  systems,'' \emph{IEEE Control Systems Magazine}, vol.~40, no.~1, pp. 26--44,
  2020.

\bibitem{albers2019online}
F.~Albers, C.~R{\"o}smann, F.~Hoffmann, and T.~Bertram, ``Online trajectory
  optimization and navigation in dynamic environments in {ROS},'' in
  \emph{Robot Operating System (ROS)}.\hskip 1em plus 0.5em minus 0.4em\relax
  Springer, 2019, pp. 241--274.

\bibitem{kumar2017search}
A.~S. Kumar, G.~Manikutty, R.~R. Bhavani, and M.~S. Couceiro, ``Search and
  rescue operations using robotic darwinian particle swarm optimization,'' in
  \emph{2017 International Conference on Advances in Computing, Communications
  and Informatics (ICACCI)}.\hskip 1em plus 0.5em minus 0.4em\relax IEEE, 2017,
  pp. 1839--1843.

\bibitem{dos2016implementing}
W.~P.~N. dos Reis and G.~S. Bastos, ``Implementing and simulating an
  alliance-based multi-robot task allocation architecture using {ROS},'' in
  \emph{Robotics}.\hskip 1em plus 0.5em minus 0.4em\relax Springer, 2016, pp.
  210--227.

\bibitem{erHos2019ros2}
E.~Er{\H{o}}s, M.~Dahl, K.~Bengtsson, A.~Hanna, and P.~Falkman, ``A {ROS}2
  based communication architecture for control in collaborative and intelligent
  automation systems,'' \emph{Procedia Manufacturing}, vol.~38, pp. 349--357,
  2019.

\bibitem{erHos2019integrated}
E.~Er{\H{o}}s, M.~Dahl, A.~Hanna, A.~Albo, P.~Falkman, and K.~Bengtsson,
  ``Integrated virtual commissioning of a {ROS}2-based collaborative and
  intelligent automation system,'' in \emph{2019 24th IEEE International
  Conference on Emerging Technologies and Factory Automation (ETFA)}.\hskip 1em
  plus 0.5em minus 0.4em\relax IEEE, 2019, pp. 407--413.

\bibitem{reke2020self}
M.~Reke, D.~Peter, J.~Schulte-Tigges, S.~Schiffer, A.~Ferrein, T.~Walter, and
  D.~Matheis, ``A self-driving car architecture in {ROS}2,'' in \emph{2020
  International SAUPEC/RobMech/PRASA Conference}.\hskip 1em plus 0.5em minus
  0.4em\relax IEEE, 2020, pp. 1--6.

\bibitem{farina2019disropt}
F.~Farina, A.~Camisa, A.~Testa, I.~Notarnicola, and G.~Notarstefano,
  ``{DISROPT}: a {P}ython framework for distributed optimization,'' \emph{arXiv
  preprint arXiv:1911.02410}, 2019.

\bibitem{koenig2004design}
N.~Koenig and A.~Howard, ``Design and use paradigms for gazebo, an open-source
  multi-robot simulator,'' in \emph{IEEE/RSJ International Conference on
  Intelligent Robots and Systems (IROS)}, vol.~3, 2004, pp. 2149--2154.

\bibitem{notarstefano2011containment}
G.~Notarstefano, M.~Egerstedt, and M.~Haque, ``Containment in leader--follower
  networks with switching communication topologies,'' \emph{Automatica},
  vol.~47, no.~5, pp. 1035--1040, 2011.

\bibitem{burger2012distributed}
M.~B{\"u}rger, G.~Notarstefano, F.~Bullo, and F.~Allg{\"o}wer, ``A distributed
  simplex algorithm for degenerate linear programs and multi-agent
  assignments,'' \emph{Automatica}, vol.~48, no.~9, pp. 2298--2304, 2012.

\bibitem{richards2007robust}
A.~Richards and J.~P. How, ``Robust distributed model predictive control,''
  \emph{International Journal of control}, vol.~80, no.~9, pp. 1517--1531,
  2007.

\bibitem{chopra2017distributed}
S.~Chopra, G.~Notarstefano, M.~Rice, and M.~Egerstedt, ``A distributed version
  of the hungarian method for multirobot assignment,'' \emph{IEEE Transactions
  on Robotics}, vol.~33, no.~4, pp. 932--947, 2017.

\bibitem{park2011smooth}
J.~J. Park and B.~Kuipers, ``A smooth control law for graceful motion of
  differential wheeled mobile robots in 2d environment,'' in \emph{2011 IEEE
  International Conference on Robotics and Automation}.\hskip 1em plus 0.5em
  minus 0.4em\relax IEEE, 2011, pp. 4896--4902.

\bibitem{mesbahi2010graph}
M.~Mesbahi and M.~Egerstedt, \emph{Graph theoretic methods in multiagent
  networks}.\hskip 1em plus 0.5em minus 0.4em\relax Princeton University Press,
  2010, vol.~33.

\end{thebibliography}

\end{document}